\documentclass[conference]{IEEEtran}
\IEEEoverridecommandlockouts
\usepackage{cite}
\usepackage{amsmath,amssymb,amsfonts}
\usepackage{algorithmic}
\usepackage{graphicx}
\usepackage{textcomp}
\usepackage{xcolor}
\usepackage{booktabs}
\usepackage{url}
\usepackage{comment}

\def\BibTeX{{\rm B\kern-.05em{\sc i\kern-.025em b}\kern-.08em
    T\kern-.1667em\lower.7ex\hbox{E}\kern-.125emX}}
\begin{document}

\title{Clinical Pathways as Safety Specifications for Physical AI in Hospital Wards}

\author{
\IEEEauthorblockN{Gabriele Franchini}
\IEEEauthorblockA{
Computer Science Dept.\\
University of Bari\\
Bari, Italy\\
g.franchini8@studenti.uniba.it
}
\and
\IEEEauthorblockN{Giulio Mallardi}
\IEEEauthorblockA{
Computer Science Dept.\\
University of Bari\\
Bari, Italy\\
giulio.mallardi@uniba.it
}
\and
\IEEEauthorblockN{Michele De Carolis}
\IEEEauthorblockA{
Computer Science Dept.\\
University of Bari\\
Bari, Italy\\
m.decarolis11@studenti.uniba.it
}
\and
\IEEEauthorblockN{Filippo Lanubile}
\IEEEauthorblockA{
Computer Science Dept.\\
University of Bari\\
Bari, Italy\\
filippo.lanubile@uniba.it
}
}

\maketitle

\begin{abstract}
Ensuring safety in Physical AI systems operating in real-world environments is a critical challenge, particularly in hospital wards where vulnerable patients, clinical staff, medical devices, and assistive robots coexist. In this paper, we reinterpret Clinical Pathways as explicit runtime safety specifications for embodied medical AI.
We propose a conceptual robotic architecture that integrates wearable sensors, smart medical devices, and assistive robotic components into a unified framework for real-time safety monitoring. At its core, a Runtime Safety Monitor (RSM) evaluates multimodal physiological and system-level signals against clinically defined constraints derived from the prescribed care process.
Rather than relying solely on statistical anomaly detection, the proposed approach combines temporal prediction, uncertainty-aware reasoning, and constraint-based verification to identify safety violations. The RSM targets three classes of events: physiological deviations from prescribed care, hardware and communication failures, and potential data tampering or misuse.
This work contributes to Safe Physical AI by operationalizing domain-specific clinical knowledge as enforceable safety constraints, bridging learning-based perception and runtime safety monitoring to assist nursing staff in real-world hospital wards.
\end{abstract}


\section{Introduction}

Physical AI systems increasingly operate in real-world environments where their decisions and actions may affect human safety. This is especially critical in hospital wards, where assistive robots, wearable sensors, and smart medical devices may operate alongside vulnerable patients and clinical staff~\cite{cruces2024socially}. In such scenarios, safety depends not only on reliable devices, but on whether the embodied system follows the intended care process despite sensor noise, communication failures, uncertainty, and misuse.

Runtime monitoring is therefore becoming central for robotic and autonomous systems deployed in uncertain environments~\cite{rahman2021runtime}. Abnormal situations are often addressed through anomaly detection, where learned models identify deviations in physiological data, sensor behavior, or clinical processes~\cite{haque2015sensor,huang2012anomaly}. While useful, anomaly scores are limited for safety-critical Physical AI: they do not necessarily explain which clinical requirement was violated, whether the deviation is clinically meaningful, or who should intervene. They are also difficult to audit when embedded in a care process involving clinicians, patients, nursing staff, and autonomous robotic components.

We argue that a more suitable abstraction already exists in healthcare: Clinical Pathways (CPs). CPs define care delivery over time, including medication schedules, measurement windows, physiological targets, and dependencies between clinical actions~\cite{schrijvers2012thecare}. Rather than treating CPs as background knowledge, this paper reinterprets them as \emph{runtime safety specifications} for embodied medical AI. The key question becomes not whether a signal is anomalous, but whether the prescribed care is being delivered safely.

Earlier work introduced an edge-node architecture for trustworthy patient home-care, including anomaly detection over deviations from Clinical Pathways~\cite{ardito2020towards}. This paper moves the problem to embodied hospital-ward systems: the objective is not to classify anomalous data streams, but to verify at runtime whether prescribed care is delivered safely by a Physical AI system.

We propose a conceptual reference architecture in which an assistive robot, wearable devices, and smart medical devices cooperate through a Runtime Safety Monitor (RSM). The RSM evaluates multimodal signals against CP-derived specifications and produces interpretable violation reports grounded in clinical constraints.

The main contributions of this paper are:
\begin{itemize}
    \item a reformulation of Clinical Pathways as runtime-verifiable safety specifications for embodied medical AI;
    \item a robotic architecture integrating wearable sensing, smart medical devices, and embodied monitoring beyond static edge-based solutions;
    \item a Runtime Safety Monitor design that combines temporal prediction, uncertainty-aware reasoning, and constraint-based verification to produce interpretable safety-violation reports.
\end{itemize}

The remainder of the paper is structured as follows. Section~\ref{sec:approach} presents the proposed approach; Section~\ref{sec:events} illustrates representative safety events; Section~\ref{sec:discussion} discusses implications, limitations, and future directions.

\section{Approach}
\label{sec:approach}

\subsection{System Overview}
\label{sec:architecture}

The proposed architecture frames clinical monitoring as a workflow in which wearable sensors, smart medical devices, and a mobile assistive robot cooperate for real-time safety monitoring. The architecture is middleware-agnostic: ROS~2 can be used as a reference implementation, but the safety-monitoring logic is not tied to a specific robotic middleware.
A platform such as TIAGo Pro\footnote{\url{https://pal-robotics.com}} can instantiate the embodied component by navigating to the patient when Clinical Pathway (CP) steps become due, supervising vital-sign acquisition, and hosting the Runtime Safety Monitor (RSM) locally. Compared with edge-node home-care monitoring~\cite{ardito2020towards}, the proposed architecture shifts from static monitoring to an embodied safety loop: the robot is not merely a data gateway, but an active sensing and interaction component that helps verify whether prescribed care is delivered safely at runtime.

\begin{figure*}[!t]
    \centering
    \includegraphics[width=\textwidth]{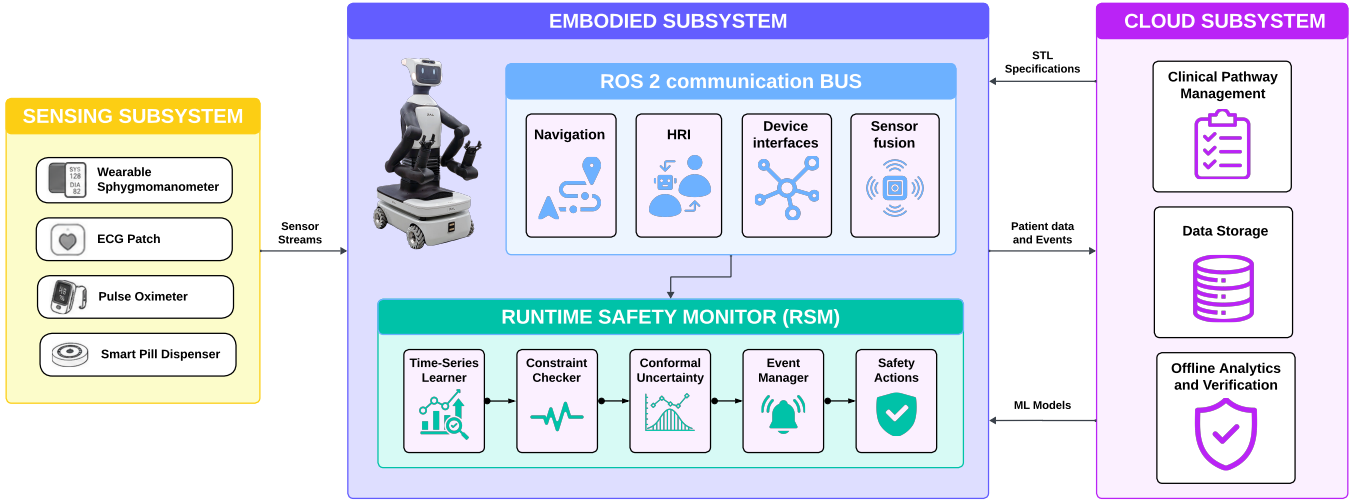}
    \caption{Robotic architecture organized into three cooperating subsystems: sensing, embodied (hosting the Runtime Safety Monitor), and edge cloud.}
    \label{fig:architecture}
\end{figure*}

The architecture is organized into three cooperating subsystems (Fig.~\ref{fig:architecture}), each with a distinct safety responsibility. 

The \textbf{sensing subsystem} provides the raw evidence for safety judgment: patient-worn Bluetooth devices (sphygmomanometer, ECG patch, pulse oximeter, smart pill dispenser) capture physiological and adherence signals prescribed by the CP, while ambient sensors supply contextual environmental signals. On-robot perception (RGB-D camera, microphone array), encapsulated within the HRI node of the embodied subsystem, provides an independent channel for presence verification and contextual grounding. This redundancy allows the system to cross-check a tampered wearable against the robot's own observations.

The \textbf{embodied subsystem}, deployed on the assistive robot, turns this evidence into action. It exposes navigation, human-robot interaction, device control, and sensor fusion as lifecycle-managed robotic middleware nodes that schedule measurement actions when CP steps become due and mediate patient interaction. Their operational states are observable to the RSM and used to disambiguate sensor faults from clinical events. 

Within this subsystem, the \textbf{Runtime Safety Monitor (RSM)} (Section~\ref{sec:rsm}) constitutes the conceptual core of the architecture: every alert goes through a single auditable verification path before escalation.

Finally, the \textbf{edge cloud subsystem} closes the loop offline, providing privacy-aware storage of feature representations, retrospective checking of the authored CP against patient history, and periodic model updates pushed back to the RSM; unlike the runtime subsystems, it is not on the safety-critical path and can tolerate intermittent connectivity.

The dataflow combines pull-based acquisition and push-based alerts. The CP schedules sensor acquisition within clinically relevant windows, while safety events propagate locally to clinicians and nursing staff without traversing the cloud. This preserves responsiveness under degraded connectivity and keeps safety reasoning at the edge.

\subsection{Runtime Safety Monitor}
\label{sec:rsm}
 \paragraph{Clinical Pathways as safety specifications}
The conceptual core of our approach is to treat Clinical Pathways not as retrospective guidelines, but as \emph{runtime safety specifications}. The reframing is enabled by the structure of CPs themselves, which already encode bounded physiological ranges, scheduled windows, ordered actions, and cross-event dependencies; what is missing is a runtime semantics, which we supply by encoding CPs in a temporal logic for continuous signals.

A CP authored by the clinician exhibits four kinds of constraints over the multivariate signal produced by the sensing subsystem: (i) \emph{range constraints} on physiological quantities (e.g.\ systolic blood pressure within a prescribed band); (ii) \emph{temporal constraints} on the timing and frequency of measurements; (iii) \emph{sequence constraints} on the order of clinical actions; (iv) \emph{cross-modal constraints} linking events across devices (e.g.\ a pill-dispenser opening must be followed, within a bounded delay, by a measurable physiological effect).

The core range, temporal, ordering, and cross-modal constraints can be encoded in Signal Temporal Logic (STL)~\cite{maler2004monitoring}, with the bounded-future operator $\Diamond_{[\Delta_1,\Delta_2]}$ capturing pharmacological response windows: after a medication intake at time $t$, the expected physiological effect must occur in some $t' \in [t+\Delta_1, t+\Delta_2]$, where $\Delta_1$ and $\Delta_2$ are the lower and upper bounds of the prescribed response delay. Range checks on scheduled measurements are enforced point-wise at the prescribed times, and event-count constraints (e.g., the daily number of measurements or intakes) are handled through windowed counters integrated in the monitor. The full STL specification associated with a CP, denoted $\varphi_{\mathrm{CP}}$, is structured as a conjunction of these constraint patterns, with parameters such as $\textsc{BP}_{\min}$, $\textsc{BP}_{\max}$, $\Delta_1$, and $\Delta_2$ instantiated per pathology by the CP templates. The translation from clinician-authored CP to STL specification is performed offline in the edge cloud and distributed to the robot at session start; by design, the robot only \emph{evaluates} specifications it cannot author or modify, keeping the authoring path under clinical governance.

\paragraph{Semantics and interpretation}
The RSM relies on two STL temporal operators with a direct clinical reading. The \emph{always} operator $\Box \, \varphi$ states that the predicate $\varphi$ must hold at every evaluation step in the monitoring horizon, so that any single failing step constitutes a violation; we use it, for instance, to assert that each scheduled blood-pressure measurement falls within the target band. The \emph{bounded eventually} operator $\Diamond_{[\Delta_1,\Delta_2]} \, \varphi$ states that, after a trigger event at time $t$, the predicate $\varphi$ must hold at least once in the response window $[t+\Delta_1, t+\Delta_2]$; we use it to assert that, for example, each medication intake must be followed by a measurable pressure decrease within the prescribed pharmacological window.

Atomic predicates are produced by a signal abstraction subsystem that lifts heterogeneous data sources to STL-compatible signals: continuous quantities (e.g., the patient's systolic blood pressure $\textsc{BP}$) are evaluated directly, while discrete events (e.g., a pill-intake event $\textsc{PILL}$) are lifted to boolean indicator signals.

We adopt a \emph{robust} STL semantics, in which each formula returns a real-valued robustness $\sigma$: positive values indicate satisfaction with a margin, negative values quantify the severity of a violation (cf.\ Table~\ref{tab:trace}). This quantitative reading allows the RSM to rank alerts by severity and exposes graded margins of safety rather than binary alarms.

Standard STL captures the main temporal and cross-modal patterns targeted by our framework, but not every CP construct: event-count constraints (e.g., requiring a fixed number of measurements over 24 hours) are handled via windowed counters in the monitor; further limits are discussed in Section~\ref{sec:discussion}. In particular, $\Diamond_{[\Delta_1,\Delta_2]}$ requires waiting for the response window to elapse before a definitive verdict can be issued, an online-monitoring challenge addressed by the pipeline below.

\paragraph{Monitor pipeline}
Given $\varphi_{\text{CP}}$, the RSM operates as a pipeline of typed stages, each consuming the output of its predecessor. A \emph{multivariate time-series learner} predicts physiological trajectories against which observed values are compared. A \emph{constraint checker} evaluates the STL formulae online, returning a real-valued robustness $\sigma$ rather than a Boolean verdict. A \emph{calibrated uncertainty subsystem}, based on conformal prediction~\cite{angelopoulos2021agentle}, wraps the predictor with prediction intervals that provide distribution-free coverage under exchangeability assumptions, so that violations are flagged only when the uncertainty envelope exits the STL-defined safe region. Between scheduled measurements, the prediction subsystem may estimate short-horizon physiological trajectories to support early warning; however, final safety verdicts remain grounded in CP-defined events and measured values.

\paragraph{Output}
At each evaluation step, the RSM emits a tuple $\langle \tau, \kappa, \sigma, \rho, \xi \rangle$ that bundles the materials produced by the three pipeline stages: the violated constraint $\tau$ and the explanation handle $\xi$ pointing to the offending sub-formula in $\varphi_{\text{CP}}$, the severity score $\sigma$ from STL robustness, and the uncertainty-aware confidence score $\rho$ from the uncertainty subsystem. Severity and confidence are deliberately kept distinct: $\sigma$ quantifies \emph{how badly} the prescribed care is violated, while $\rho$ indicates \emph{how confidently} a violation can be claimed. These are two orthogonal axes that a single anomaly score would conflate. The field $\kappa$ classifies the alert into one of the three event categories of Section~\ref{sec:events}, letting a single monitor handle genuine clinical deviations, infrastructure failures, and adversarial tampering through the same machinery.

This design supports \emph{intrinsic} explainability: each alert points, via $\xi$, to the violated sub-formula of $\varphi_{\text{CP}}$, without post-hoc attribution. An event manager then maps each tuple to the appropriate safety action, such as notifying the clinician, locking the dispenser, or aborting a procedure.

\section{Safety Events and Use Case}
\label{sec:events}

\subsection{A Taxonomy of Safety Events}

The constraint structure of Section~\ref{sec:rsm} induces a natural taxonomy of three classes, each corresponding to a distinct mode of violation of $\varphi_{\text{CP}}$.

\textbf{(C1) Physiological deviations.}
A measured vital sign falls outside the range or temporal envelope prescribed by the CP, violating range or cross-modal sub-formulae. Typical instances include hypertensive spikes, post-medication trajectories that fail to converge to the expected band within the prescribed delay, or missed measurements at scheduled times.

\textbf{(C2) System and embodiment failures.}
A violation is induced not by the patient but by the monitoring infrastructure itself. This class subsumes the hardware faults considered in earlier work~\cite{ardito2020towards}, such as battery depletion, probe detachment, and Bluetooth errors, and extends them to the robot platform, including localization drift, manipulation failures during pillbox interaction, and missed deadlines on safety-critical topics. By correlating predictor residuals with the lifecycle state of the involved robotic middleware nodes, the RSM ensures that a faulty sensor cannot masquerade as a clinical event.

\textbf{(C3) Adversarial tampering.}
A violation arises from intentional manipulation of the data stream or the embodied agent. Beyond classical data-injection on the wearable bus, the embodied setting introduces new attack surfaces: spoofing of patient presence to the robot's perception stack, or physical manipulation of the smart pill dispenser to forge adherence events. Such events typically violate cross-modal constraints, for example a \textsc{PILL} event without the expected downstream physiological response, and surface as low confidence on otherwise plausible-looking signals.

A single monitor handles all three classes: what differs is which sub-formula of $\varphi_{\text{CP}}$ is violated and how the uncertainty-aware confidence score $\rho$ distributes between predictor and constraint checker. This unification is, in our view, the practical payoff of the safety-specification reframing.

\subsection{Running Example: Hypertension Monitoring}
\label{sec:usecase}

We illustrate the three classes through an embodied hypertension monitoring scenario. The CP prescribes two antihypertensive pills (07:00, 21:00) with an expected blood-pressure response within 1--3 hours, five daily blood-pressure measurements with a target band $\textsc{BP} \in [110,140]$~mmHg, and counter constraints requiring five measurements and two intakes over 24 hours. At each scheduled step, the assistive robot navigates to the patient, verifies presence via its RGB-D stack, and triggers the wearable sphygmomanometer.

Table~\ref{tab:trace} reports an illustrative one-day trace, in which one event of each class arises. The \textbf{C1} deviation at 08:55 is a genuine clinical event: predictor, constraint checker, and conformal layer all agree, yielding strongly negative $\sigma$ and high $\rho$. The \textbf{C2} failure at 15:00 is detected by correlating the truncated measurement window with the sphygmomanometer node transitioning to \textsc{inactive} mid-task: the RSM attributes the violation to hardware rather than to the patient. The \textbf{C3} tampering at 21:02, a \textsc{PILL} event without a downstream pressure decrease, looks plausible to the predictor in isolation, but the constraint checker fires on the cross-modal and dosage sub-formulae and the confidence score drops, prompting the robot to escalate to the clinician and lock the dispenser.

\begin{table}[t]
\centering
\caption{Illustrative one-day RSM trace, not an empirical evaluation. $\sigma$: STL robustness (negative = violation). $\rho \in [0,1]$: confidence (higher is more confident). $\tau$: violated sub-formula ($\varphi_{\text{BP}}$: BP range; $\varphi_{\text{HW}}$: hardware-liveness; $\varphi_{\text{DOSE}}$: dosage counter).}

\label{tab:trace}
\small
\begin{tabular}{@{}llrrll@{}}
\toprule
Time   & Event                          & $\sigma$ & $\rho$ & $\tau$                  & Class \\
\midrule
07:00  & \textsc{pill} (dispenser open) & $+18$    & 0.97   & ---                     & ok    \\
08:55  & \textsc{bp}=152~mmHg           & $-12$    & 0.94   & $\varphi_{\text{BP}}$   & C1    \\
15:00  & \textsc{bp\_meas} (9~s window) & $-21$    & 0.71   & $\varphi_{\text{HW}}$   & C2    \\
21:00  & \textsc{pill} (scheduled)      & $+15$    & 0.96   & ---                     & ok    \\
21:02  & \textsc{pill} (unscheduled)    & $-8$     & 0.42   & $\varphi_{\text{DOSE}}$ & C3    \\
\bottomrule
\end{tabular}
\end{table}

\section{Discussion}
\label{sec:discussion}

The architecture instantiates a broader shift in runtime safety for embodied AI: from data-driven anomaly detection, which treats deviations as statistical outliers, to specification-based monitoring, where deviations are interpreted against an explicit, human-authored model of intended behavior. In clinical environments this shift is particularly natural, since the specification already exists, the Clinical Pathway, and what is missing is the machinery to turn it into a runtime artifact: the CP becomes a verifiable safety contract, the time-series learner provides robust evaluation under noise, and the calibrated uncertainty subsystem turns the resulting verdicts into confidence-aware alerts.

Three aspects connect this work to the trustworthy-AI agenda. \emph{Verifiability}: every alert is anchored to a sub-formula of the clinician's prescription, not to a latent representation. \emph{Calibration}: the conformal subsystem separates \emph{what} is happening from \emph{how confidently} we can claim it, a prerequisite for any meaningful confidence score in safety-critical contexts. \emph{Embodiment as a safety asset}: the robot's active sensing yields redundancy a passive network cannot, allowing a tampered wearable to be cross-checked against on-board perception.

Three limitations remain open. \emph{(i) Specification authoring.} Translating a clinician-authored CP into STL presupposes either a template library or a clinician-facing authoring tool; the parameter values instantiating these templates must be grounded in clinical guidelines and pharmacokinetic data, and STL expressiveness ultimately bounds the safety properties one can state. \emph{(ii) Distribution shift.} Conformal coverage guarantees rely on exchangeability assumptions that physiological signals routinely violate around clinical events; adaptive variants mitigate but do not eliminate the issue. \emph{(iii) Clinical validation and implementation status.} The pseudo-trace of Section~\ref{sec:usecase} establishes feasibility, not efficacy, and the architecture is not deployed; studies with real devices, patients, and clinicians are required before any deployment claim.

Several extensions follow naturally from the architecture. Patient-specific calibration could adapt CP parameters to individual baselines while preserving the logical structure of the specification. Offline retrospective checking of authored CPs against patient history could further support clinical governance. Future empirical studies should also evaluate how clinicians and nursing staff interpret RSM outputs, especially when severity and confidence diverge.

\section{Conclusion}
\label{sec:conclusion}

We have argued that Clinical Pathways, long treated as care-flow artifacts, can be operationalized as runtime safety specifications for embodied AI in hospital wards. We sketched a clinical robotic architecture in which an assistive robot hosts a Runtime Safety Monitor that evaluates CP-derived constraints against multimodal sensing streams under uncertainty. The reframing replaces \emph{is this signal anomalous?} with \emph{is the prescribed care being delivered safely?}, making every alert traceable to a clinically meaningful constraint. This provides a first step towards Physical AI systems whose trustworthiness is not merely asserted, but checked at runtime against the specification that defines their purpose.

\section*{Acknowledgment}
This research was co-funded by the Complementary National Plan PNC-I.1 (``DARE'', PNC0000002, CUP: B53C22006420001).

\bibliographystyle{IEEEtran}
\bibliography{bib}
\end{document}